\documentclass[]{cit_lab_mfr}

\usepackage{amssymb}

\usepackage{amsthm}
\usepackage{amsmath}
\usepackage{mathtools}
\usepackage{enumitem}
\usepackage{float}
\usepackage{adjustbox}
\usepackage{array}
\usepackage{pifont}

\graphicspath{{images/}{figures/}}

\newcommand{\methodname}{\texorpdfstring{Fysiverse-3D-SimReady}{Fysiverse-3D-alpha}}

\providecommand{\Description}[1]{}
\title{\methodname{} Technical Report: Agentic Physical \\Simulation for Pragmatic 3D World Reconstruction}

\author{%
\parbox{\textwidth}{\centering
Lintao Wang, Mingyang Sun, Yang Liu, Dingkang Yang$^{\S,\dagger}$, Lihua Zhang$^{\S}$
}}

\affiliation{%
\parbox{\textwidth}{\centering\small
Physical Superintelligence Lab, Fysics AI \\[1mm]
College of Intelligent Robotics and Advanced Manufacturing, Fudan University\\[1mm]
Multimedia Laboratory (MMLab), The Chinese University of Hong Kong\\[1mm]
College of Electronic and Information Engineering, Tongji University\\[1mm]
}}

\contribution[\dagger]{Project lead}
\contribution[\S]{Corresponding author}

\abstract{
Agentic recognition requires visual perception to move beyond static scene understanding and produce structured scene representations that support the perception--reasoning--action loop. Existing single-image 3D generation methods, however, mainly produce visually plausible object assets rather than simulation-ready scene states. When independently generated meshes are composed in a shared space, they may fail to align with the input camera, violate gravity, interpenetrate nearby objects, or become unstable under physics simulation. We present \methodname{}, a grounded refinement framework for reconstructing simulation-ready multi-object scenes from a single RGB image with instance and ground prompts. The method places generated object meshes into a shared gravity-aligned scene, refines their camera-space poses through differentiable rendering, and corrects scene-level supports and contacts for stable physical execution. The scene is then used by an agentic simulation workflow, which converts a scene-specific task goal into an executable physics rollout rendered from the original camera view. Experiments show that \methodname{} improves input-view alignment, contact plausibility, and physical stability over existing single-image reconstruction and scene generation baselines, while enabling goal-conditioned physical interactions from a single image.
}
\date{\today}
\checkdata[Corresponding]{ \url{dicken@fyscis.ai}, \url{lihuazhang@fudan.edu.cn}}
\checkdata[Page]{\url{https://fysics-ai.github.io/Fysiverse-3D-project-page/}}
\checkdata[Github]{\url{https://github.com/Fysics-AI/Fysiverse-3D-SimReady}}

\begin{document}
\maketitle

\section{Introduction}
\label{sec:introduction}

Agentic AI~\cite{yang2025medaide,yang2026toward,qian2026spatialguard,yang2025improving,han2026omnifysics} is expanding pattern recognition beyond static visual interpretation toward systems that use structured scene representations to interpret task-specific physical goals and invoke external simulation tools~\cite{kang2026mtlq,wang2026ced,fang2026belief,pu2025leveraging}. For agentic physical simulation, such representations must be not only visually grounded but also geometrically coherent and suitable for stable physics execution~\cite{zhu20265,zhu2026omnirag,gao20253d}. Single-image 3D generation provides a promising basis for constructing these representations because it can recover visually plausible object meshes from image crops. However, independently generated assets do not directly constitute a simulation-ready scene. They must first be placed in a shared camera and gravity frame, assigned consistent support and contact relationships, and equipped with physical representations that can be reliably consumed by a simulator. Two lines of recent progress make this problem increasingly important and tractable. On the one hand, image-based 3D models can recover increasingly detailed object geometry and materials from sparse visual input. Feedforward reconstruction models make this possible from a single crop \cite{hong2023lrm,tochilkin2024triposr}, while recent mesh, latent, and textured asset generators further improve geometric detail and appearance \cite{wu2024unique3d,xiang2025trellis,hunyuan3d2025}. On the other hand, real-to-sim systems have shown that simulated scenes can support policy evaluation, data generation, and interactive manipulation when their geometry and dynamics are sufficiently grounded in the real world \cite{torne2024rialto,han2025re3sim,jain2025polaris,li2024simmanip}. Despite this progress, a mesh generated from a single image remains defined in its own local frame and carries inherent ambiguities in scale, pose, and contact. When multiple assets are assembled into a shared scene, these ambiguities accumulate into scene-level failures, including inconsistent camera alignment, unreliable gravity, incorrect support relationships, and unstable physical behavior.

The missing layer is scene-level grounding.
For a single image, the available evidence is mostly projective:
object regions, visible boundaries, and occlusions.
The quantities that matter for simulation are different: gravity,
contact, collision geometry, and scene stability.
Recent systems for simulation-ready scene generation and embodied
environments emphasize this need for physical structure
\cite{pfaff2026scenesmith,wang2025tabletopgen,xia2026sage}, while
practical simulation pipelines often rely on simplified collision
proxies and physics engines rather than the raw visual mesh
\cite{wei2022coacd,xiang2020sapien,todorov2012mujoco}.
Our goal is to bridge these requirements in the single image setting.
Direct composition exposes a gap between asset generation and scene
generation.
A mesh can look reasonable in isolation while floating above a table,
penetrating another object, rotating around the wrong gravity axis, or
projecting to the wrong object region from the input view.
These errors are not cosmetic.
They determine whether the generated scene can be rendered back into the
source image, whether physical contact is meaningful, and whether a
simulated video remains stable after the first few frames.

We address this gap by treating single image scene generation as a
grounded refinement problem.
Figure~\ref{fig:method-overview} illustrates the resulting pipeline, from
prompted single image inputs to consistent initialization,
view alignment, contact-aware scene correction, and
goal-conditioned simulation rendering.
Instead of asking a generator to solve all visual, geometric, and
physical constraints in one pass, we initialize the scene with
generated meshes and then refine the result with constraints that come
from the input image and the spatial positions of the generated objects.
The method keeps the early stage deliberately permissive: generated
assets only need to provide editable geometry and a reasonable visual
prior.
Scene-level constraints are introduced later, when the camera, gravity
direction, and contacts can be checked against the image and
corrected without forcing the generator itself to solve a fully physical
scene from a single view.
Overall, this paper makes the following contributions:
\begin{itemize}
  \item We propose \methodname{} (F3SR), a unified framework that connects
  single image scene reconstruction with agentic physical simulation,
  turning a static RGB image into a simulation-ready scene that can be
  driven by scene-specific task goals.
  \item We introduce a simulation-ready scene reconstruction pipeline that
  grounds independently generated assets in a shared gravity frame and the
  original camera, while correcting supports, contacts, and penetrations so
  the reconstructed scene can be used for physics simulation.
  \item We develop an agentic workflow for physical simulation. It starts from
  the refined scene and a scene-specific task goal to build an executable
  rollout rendered from the original view.
\end{itemize}

\section{Related Work}
\label{sec:related-work}

\subsection{Agentic Workflows and Physical Simulation}

Language-based agents have become a practical interface between high-level goals and executable tools. In robotics, SayCan grounds language-model planning with affordance scores from available robot skills, while ProgPrompt converts natural-language instructions into situated program plans \cite{ahn2022saycan,singh2022progprompt}. Inner Monologue further incorporates environment feedback, scene descriptions, and success detection into the planning process \cite{huang2022innermonologue}. Agentic workflows have also been introduced into 3D environment construction. SceneSmith employs iterative generation and feasibility checking to construct simulation-ready indoor scenes, while SAGE uses semantic, visual, and physical critics to refine embodied AI environments \cite{pfaff2026scenesmith,xia2026sage}. These studies demonstrate the value of decomposing complex objectives, maintaining structured intermediate states, and validating tool inputs before execution. 
\methodname{} adopts this principle for physical simulation rather than robot action generation. Its agentic workflow interprets a scene-specific physical goal, assigns object roles, proposes physical parameters and initial conditions, validates the resulting plan, and invokes deterministic simulation and rendering tools over the reconstructed scene. 
\subsection{3D Generation and Scene Construction}

Single-image 3D generation provides the object-level foundation for the proposed setting. Feedforward models such as LRM and TripoSR learn large-scale 3D priors and map an image crop to an object representation at practical inference speed \cite{hong2023lrm,tochilkin2024triposr}. Image-conditioned diffusion methods improve object generation by predicting novel views and enforcing multi-view consistency \cite{shi2023mvdream,long2023wonder3d}. Other systems focus on efficient textured-mesh or Gaussian reconstruction from sparse generated views \cite{wu2024unique3d,wang2024crm,xu2024instantmesh}, while compact structured latents and stronger texture models further improve asset fidelity \cite{xiang2025trellis,hunyuan3d2025}.
Scene-level methods extend this capability from isolated assets to multi-object environments. SAM3D, MIDI, and SceneGen estimate object geometry, pose, or layout from a single image \cite{sam3dteam2025sam3d3dfyimages,huang2025midi,meng2026scenegen}. Modular world-building systems such as Flash Sculptor further emphasize object decomposition and compositional assembly \cite{hu2025flashsculptor}. These methods substantially improve object fidelity and visual scene composition. However, independently generated assets may still carry incompatible scales, poses, coordinate frames, and contact assumptions. Consequently, visual fidelity and layout consistency alone do not ensure that the resulting scene can be stably executed in a physics simulator.

\subsection{Simulation-Ready Scene Generation and Simulation Platforms}

Simulation-ready reconstruction introduces physical requirements beyond visual scene composition. TabletopGen constructs interactive tabletop scenes from text or a single image by estimating instance-level pose and scale before simulator assembly \cite{wang2025tabletopgen}. PAT3D incorporates physics-aware constraints into text-based 3D scene generation \cite{lin2025pat3d}. SPARCS jointly optimizes convex object shapes and poses to avoid penetration while satisfying force-balance and friction constraints, although it assumes RGB-D input \cite{huang2026sparcs}. These methods highlight the importance of gravity consistency, stable support, non-penetrating contact, and physically valid object placement.
Physical execution also depends on appropriate simulation abstractions and reliable platforms. Embodied AI environments such as Habitat and iGibson demonstrate the value of structured interactive scenes for navigation, manipulation, and everyday activities \cite{savva2019habitat,li2022igibson}. Real-to-sim systems such as RialTo, Re3Sim, PolaRiS, and SIMPLER use richer observations, demonstrations, or real-world evaluation to construct simulation environments for manipulation learning and policy assessment \cite{torne2024rialto,han2025re3sim,jain2025polaris,li2024simmanip}. At the execution level, SAPIEN and MuJoCo provide controllable physics rollouts, while CoACD decomposes complex meshes into collision-aware convex parts that are more suitable for stable contact simulation \cite{xiang2020sapien,todorov2012mujoco,wei2022coacd}. These research directions establish complementary components of simulation-ready scene construction, but they address different stages under different input assumptions. The proposed method brings these requirements together in the single-RGB setting by combining original-view grounding, relation-aware contact correction, collision proxies, passive settling, and downstream agentic physical simulation within a unified pipeline.

\section{Problem Formulation}
\label{sec:problem}

Our setting begins with a photograph and asks for more than a static 3D
reconstruction: the recovered scene should remain tied to the photograph
while also being usable in a physics simulator.
The input is a single RGB image $I$, a set of instance masks
$\mathcal{M}=\{m_i\}_{i=1}^{N}$ obtained from object prompts, and a
ground mask $m_g$ obtained from a separate ground prompt:
\begin{equation}
  \mathcal{X} = (I,\mathcal{M},m_g).
\end{equation}
Here, the ground prompt refers to a separate scene-level prompt that
marks the visible support surface, such as a floor, tabletop, countertop,
or ground. It provides evidence for gravity estimation and the
simulator ground plane, but it is not treated as another foreground
object to reconstruct.
The masks provide visible evidence for the objects and the ground, but
they are only the anchors for scene construction.
A simulation-ready output must additionally assign depth, scale, contact
structure, and physical representations in a common 3D frame.

The desired output is a scene state, denoted by $\mathcal{Y}$, that can
be both rendered from the input view and advanced by a physics simulator.
We write this state as:
\begin{equation}
  \mathcal{Y} =
  \left(C,\{M_i,T_i,s_i,P_i\}_{i=1}^{N},\tau\right),
  \label{eq:problem-output}
\end{equation}
where $C$ is the camera corresponding to the input view, $M_i$ is the
textured visual mesh of object $i$, $T_i$ and $s_i$ are its pose and
scale in a shared world frame, $P_i$ is a collision proxy used by the
physics simulator, and $\tau$ denotes the pose sequence produced by a
physics rollout.
A collision proxy is a simplified geometric representation of the object
used for contact and collision computation; it need not match the visual
mesh exactly.
This representation separates appearance from simulation: the visual
mesh should preserve the generated asset quality, while the collision
proxy should make contact-aware physical motion tractable.

We call a scene simulation-ready if it satisfies three high-level
requirements.
First, it should preserve input image grounding: when rendered from $C$,
the objects are expected to remain well aligned with the observed instance masks.
Second, the objects should live in a coherent world frame with a
consistent gravity direction.
Third, the scene should be physically usable. Contacts should be
plausible, object interpenetration should be avoided, and the scene
should not collapse or drift severely after settling.
Let $\Pi_C(\mathcal{Y})$ denote the rendering of the scene state from
camera $C$, reduced to the same mask domain as $\mathcal{M}$, and let
$\Omega_{\mathrm{ready}}$ denote the set of scenes that satisfy the
gravity, contact, and stability requirements above.
The task can be summarized as:
\begin{equation}
  \mathcal{Y}^{\star} = F(\mathcal{X}), \qquad
  \Pi_C(\mathcal{Y}^{\star}) \approx \mathcal{M}, \qquad
  \mathcal{Y}^{\star}\in\Omega_{\mathrm{ready}}.
  \label{eq:abstract-problem}
\end{equation}
Here $F$ denotes the overall scene construction process.
The middle condition requires the output to remain faithful to the source
image, while the final condition requires the same scene state to be
usable by a physics simulator.

The formulation places the visual and physical requirements on the same
artifact.
A generated object collection becomes a scene only after its members
share the camera, world frame, and simulation state needed for
original view rendering and simulator-side settling.

\section{Methodology}
\label{sec:method}

F3SR reconstructs a simulation-ready scene by separating object
generation from scene-level refinement and goal-conditioned
simulation.
Given a prompted RGB image, the system first obtains one editable mesh for
each visible instance and then places all meshes into a shared scene where
camera alignment, gravity, support, contact, and simulation behavior can be
refined jointly.
A key design choice is to preserve object identity throughout the pipeline.
The same mask is used to connect each instance with its generated visual
mesh, semantic description, support relations, collision proxy, simulation
actor, and original view diagnostic render.
This identity tracking prevents later language and physics modules from
changing the object set, and makes each geometric or physical correction
traceable to an observed image region.

\begin{figure}[!t]
  \centering
  \includegraphics[width=\linewidth,height=0.76\textheight,keepaspectratio]{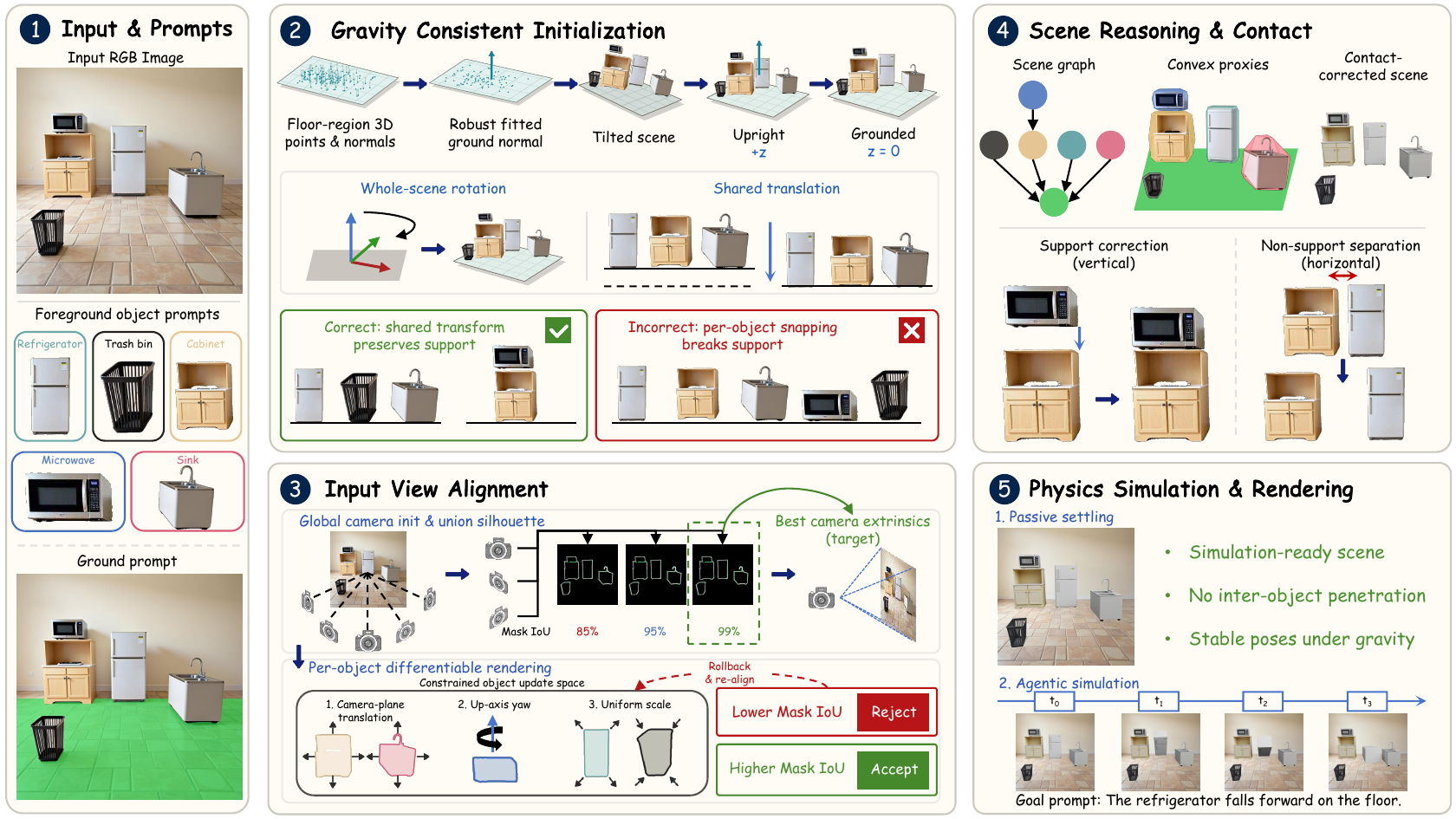}
  \caption{Overview of the \methodname{}. The pipeline builds object assets from a
  prompted RGB image, aligns them to gravity and the input view, corrects
  contacts with collision proxies, and renders physical rollouts from the original
  view.}
  \Description{A five-stage method overview. The input image provides object
  prompts and a ground prompt. Gravity-consistent initialization estimates a
  ground normal, applies a shared whole-scene rotation, and grounds the scene
  without per-object snapping. Input view alignment first optimizes the global
  camera with a union silhouette and then refines per-object translation,
  yaw, and scale with overlap-based acceptance or rollback. Scene reasoning builds
  a graph and convex proxies for contact correction. The final stage shows
  passive settling evidence and an agentic goal-conditioned simulation
  rollout rendered from the original view.}
  \label{fig:method-overview}
\end{figure}

The method is organized around three forms of grounding.
The first places generated meshes into a gravity-consistent world frame.
The second aligns this world to the input view using mask-based
differentiable rendering.
The third uses mask-scoped VLM scene graph reasoning, relation-aware
contact correction, and physical settling to turn a visually aligned
static scene into a simulation-ready scene.
This scene graph call provides semantics and support priors for
reconstruction, while the explicit Planner, Reasoner, and Builder workflow is
reserved for downstream goal-conditioned simulation.
Figure~\ref{fig:method-overview} summarizes the full flow while expanding
the key operations in gravity initialization, input view alignment,
scene-level contact correction, and physics rollout.
The goal-conditioned simulation stage assigns specialized roles to language
agents, proposes physical parameters, validates the resulting simulation plan
using deterministic tools, and renders the executed rollout from the original
camera view.

\subsection{Gravity Consistent Scene Initialization}

Scene initialization establishes a common physically meaningful frame for
independently generated objects before view alignment and contact
correction.
The input prompts play two different roles in this stage.
The instance masks $\{m_i\}$ define the foreground objects for 3D asset
generation and keep each object tied to its image region.
In practice, such masks can be produced by promptable segmentation and
open vocabulary detection models
\cite{kirillov2023segment,liu2023groundingdino}.
The ground mask $m_g$ is not treated as another object to be generated.
Instead, it provides a scene-level geometric cue for determining the
gravity direction and the scene-level plane.

Each instance mask is processed independently by a single image 3D asset
generator to obtain an editable textured object mesh.
This stage is only required to produce object assets that can be refined
later; it is not expected to solve scene layout, camera agreement, or
contact relationships in one step.
Scene-level constraints are therefore introduced after all objects are
placed in the same world coordinate system.
To estimate the up direction of this frame, we combine the ground prompt
with monocular geometry prediction, following the broad use of dense depth
and surface normal predictors for single image geometry
\cite{ranftl2021dpt,eftekhar2021omnidata,yin2023metric3d,yang2024depthanythingv2}.
Let $\mathbf p(u)$ and $\mathbf n(u)$ be the dense 3D point and normal
predicted at image pixel $u$, and let $\mathcal G$ be the set of valid
ground candidate pixels.
When a ground mask is available, $\mathcal G$ is formed from valid pixels
inside $m_g$; otherwise it is estimated from the lower image region after
foreground objects are removed.
The ground plane normal is obtained by fitting these candidate points:
\begin{equation}
  \hat{\mathbf n}_{p}
  =
  \underset{\|\mathbf n\|_2=1}{\operatorname{arg\,min}}
  \sum_{u\in\mathcal G}
  \left(\mathbf n^\top(\mathbf p(u)-\bar{\mathbf p})\right)^2 ,
  \label{eq:ground-plane}
\end{equation}
where $\bar{\mathbf p}$ is the median sampled point.
When pixel normals are reliable, we further use normals that agree with
the fitted plane as a local vote.
The final ground normal is written as:
\begin{equation}
  \hat{\mathbf n}_{g}
  =
  \operatorname{normalize}
  \left((1-\lambda)\cdot\hat{\mathbf n}_{p}
  + \lambda\cdot\operatorname{median}_{u\in\mathcal G'}\mathbf n(u)\right),
  \label{eq:ground-normal}
\end{equation}
where $\mathcal G'$ keeps pixel normals whose directions agree with the
fitted plane.
Because a fitted plane normal is sign ambiguous, we first orient it toward
the expected up direction in the source frame and, after coordinate
conversion, flip it when necessary to agree with the target up axis.

After estimating the ground direction, we express it in the simulation
coordinate frame and denote it by $\tilde{\mathbf n}_g$.
Let $\mathbf g=(0,0,1)^\top$ be the target up direction in this frame.
The upright rotation $\mathbf R_g$ is defined as the shortest rotation
that aligns $\tilde{\mathbf n}_g$ with $\mathbf g$.
This rotation is applied to the generated scene as a whole rather than to
each object independently:
\begin{equation}
  M_i^{u} = \mathbf R_g \cdot M_i,\qquad
  T_i^{u} = \mathcal T(\mathbf R_g,\mathbf 0)\cdot T_i .
  \label{eq:upright}
\end{equation}
Here $\mathcal T(\mathbf R,\mathbf t)$ denotes a rigid transform with
rotation $\mathbf R$ and translation $\mathbf t$.
The upright scene then receives one shared translation along $\mathbf g$
so that its bottom support layer rests on the simulation ground plane
$z=0$.
This global grounding does not snap each object to the floor separately,
and therefore preserves the relative height relationships already present
in the generated scene.

\subsection{Input View Alignment}

The gravity-consistent scene still carries errors from object-level
single image generation:
the camera can be slightly misoriented, object depths can drift, and
generated shapes may not cover their source masks from the input view.
Our framework resolves these errors in the input view with a staged
differentiable renderer, building on differentiable rasterization and 3D
rendering toolkits
\cite{kato2018neuralrenderer,liu2019softras,chen2019dibr,ravi2020pytorch3d,laine2020nvdiffrast}.
The stage order is important: global viewpoint error is assigned to the
camera before object-level pose changes are allowed.
The input view branch in Figure~\ref{fig:method-overview} visualizes this
progression from global camera correction to constrained object correction
with a consistent contour encoding.
We first optimize only the camera extrinsics while keeping the monocular
intrinsics fixed.
Let $S(C,\mathcal Y)$ be the differentiably rendered foreground
silhouette of all objects from camera $C$, and let $m_{\cup}=\bigvee_i
m_i$ be the union of instance masks.
The camera update minimizes:
\begin{equation}
    \mathcal L_{\mathrm{cam}}
    =
    \|S(C,\mathcal Y)-m_{\cup}\|_1
    +
    \lambda_{\mathrm{dt}}
    \operatorname{mean}\!\left(S(C,\mathcal Y)\odot D_{\bar m_{\cup}}\right)
    +
    \lambda_r\|\boldsymbol\omega\|_2^2
    +
    \lambda_t\|\mathbf t\|_2^2 .
\end{equation}
where $(\boldsymbol\omega,\mathbf t)$ parameterizes a small extrinsic
update around the initial camera, $D_{\bar m_{\cup}}$ is the normalized
distance transform outside the target mask, and $\odot$ denotes elementwise
multiplication.
The distance term penalizes rendered silhouette mass that leaks away from
the observed foreground.
The optimized camera is accepted only when its mIoU, the mean intersection over union between rendered and input instance masks, does not drop
below that of the initial camera; otherwise the original estimate is kept.
Object refinement starts from the accepted camera.
Each object is optimized against its own label mask, and objects can be
processed in parallel because their updates are applied independently
before the contact stage.
The update for object $i$ is deliberately low dimensional:
\begin{equation}
    \Delta T_i
    =
    \operatorname{Trans}(\mathbf B_C\boldsymbol\alpha_i)
    \cdot
    \operatorname{Rot}_{z}(\theta_i)
    \cdot
    \operatorname{Scale}(1+\sigma_i).
  \label{eq:object-delta}
\end{equation}
where $\mathbf B_C$ is the camera plane translation basis, $\theta_i$ is
a yaw rotation around the world $+Z$ axis, and $\sigma_i$
is a uniform scale change; both yaw and scale are applied about the object
bounding box center.
The default constraints are bounded camera plane translation, bounded
yaw, and bounded scale.
These restrictions remove the most unstable single view degrees of
freedom: arbitrary depth translation, arbitrary 3D rotation, and
anisotropic scale can all improve projection matching while damaging contact
geometry.
For each object, the rendered silhouette $S_i$ is optimized with mask,
outside distance, center, and area terms:
\begin{equation}
  \begin{aligned}
  \mathcal L_i
  =&\,
  \|S_i-m_i\|_1
  +
  \lambda_{\mathrm{dt}}\operatorname{mean}(S_i\odot D_{\bar m_i})
  \\
  &+
  \lambda_c\|\boldsymbol\mu(S_i)-\boldsymbol\mu(m_i)\|_2^2
  +
  \lambda_a
  \left(\log\frac{|S_i|}{|m_i|}\right)^2
  +
  \mathcal L_{\mathrm{prior}} .
  \end{aligned}
  \label{eq:object-loss}
\end{equation}
The center and area terms make the loss less sensitive to accidental
boundary matches.
$\mathcal L_{\mathrm{prior}}$ contains quadratic penalties on translation,
yaw, and scale updates, keeping the solution close to the initial
estimate.
In practice, translation begins with a coarse bounded seed search in the
camera plane, yaw and scale each evaluate a small set of scalar seeds,
and the three variables are optimized in the order translation, yaw,
scale.
After every phase, a mask intersection over union (mIoU) guard checks whether
the new silhouette has degraded beyond a small tolerance; a failed phase is
rolled back without discarding earlier accepted phases.

\subsection{Scene Graph Guided Physical Refinement}

Input view alignment produces a scene that is aligned with the observed view, but
depth and contact errors can remain along directions that are weakly observed.
To correct these errors without changing the object set, F3SR builds
a mask-scoped scene graph over the detected instances. The graph records object
semantics, pairwise relations, and support relations, and is constrained to the
segmented objects so that language reasoning cannot introduce new physical
bodies. These relations are used only as priors for geometric correction; the
actual contact tests are performed with explicit collision proxies.
Support relations determine how penetration is resolved. Each support relation
identifies an upper object and the object that supports it. Candidate supports
are filtered into a forest so that each child has at most one support parent
and cycles are rejected. Supported objects are corrected along the gravity
axis, while unrelated overlaps are separated horizontally. This distinction
preserves support relations implied by the image instead of pushing stacked
objects apart in the image plane.

Collision geometry is built from simplified proxies rather than
visual meshes. For object $i$, the proxy is a set of convex parts
$P_i=\{P_{ik}\}_{k=1}^{K_i}$ obtained by approximate convex decomposition.
We use bounding boxes as a broad phase filter and test candidate convex part
pairs with the separating axis theorem:
\begin{equation}
    \operatorname{collide}(i,j)
    =
    \bigvee_{k,\ell}
    \operatorname{SAT}(P_{ik},P_{j\ell}).
  \label{eq:convex-collision}
\end{equation}
Here $\operatorname{SAT}$ denotes a separating axis test between two convex parts.
This proxy is also the geometry exported to the physics engine, while the
high detail visual mesh remains available for rendering.
If the convex proxies of an upper object and its support collide, the
system searches for the smallest upward displacement that makes the pair no
longer collide and then adds a small support gap:
\begin{equation}
    \delta_z^{ij}
    =
    \min_{\delta\ge 0}
    \left\{
    \delta \mid
    \neg\operatorname{collide}(P_i^\delta,P_j)
    \right\}
    +\epsilon_g,
    \qquad
    P_i^\delta=\mathcal T(\mathbf I,\delta\mathbf g)\cdot P_i .
  \label{eq:support-correction}
\end{equation}
Here $P_i^\delta$ denotes the convex proxy of the upper object after applying
an upward translation $\delta\mathbf g$.
The displacement is applied to the entire supported subtree so that
objects stacked on the corrected object move together. Non support pairs are
handled separately: bounding box overlap proposes candidate pairs, convex SAT
confirms true penetration, and the horizontal component of the minimum
translation direction separates the pair. Support pairs are excluded from this
horizontal pass.

\subsection{Agentic Goal Conditioned Simulation}

\begin{figure}[!t]
  \centering
  \includegraphics[width=\linewidth,height=0.72\textheight,keepaspectratio]{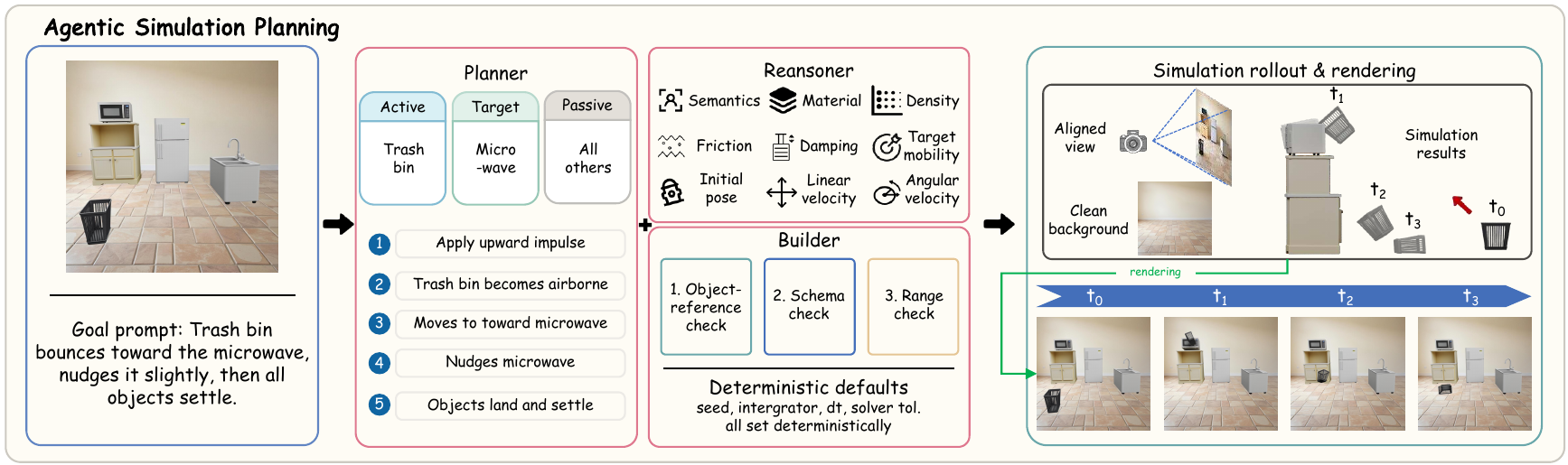}
  \caption{Agentic simulation planning. Planner, Reasoner, and Builder convert a
  simulation-ready scene and a goal into a validated simulation rollout rendered
  from the original view.}
  \Description{A method diagram for goal conditioned simulator assistance. A
  kitchen scene and a goal prompt ask a trash bin to bounce toward a microwave
  and nudge it. The Planner marks the trash bin as active, the microwave as
  target, all other objects as passive, and lists causal phases. The Reasoner
  estimates semantics, material, density, friction, damping, target mobility,
  initial pose, and velocities. The Builder checks object references and
  parameter ranges, fills deterministic defaults, and connects the plan to
  aligned view rollout and clean background rendered time sequence results.}
  \label{fig:agentic-planning}
\end{figure}

After contact correction, the scene is ready for goal conditioned simulation.
The agentic module receives the refined scene, preserved object identities,
collision proxies, the ground plane, the original camera, and a user specified
physical goal. The goal is not treated as a request to synthesize an animation
directly; instead, it is translated into an executable physics rollout over the
reconstructed scene, as shown in Figure~\ref{fig:agentic-planning}.

The module organizes this process into Planner, Reasoner, and Builder roles.
The Planner grounds the goal to existing scene objects, assigns active, target,
and passive actors, and decomposes the interaction into causal phases. The
Reasoner estimates physical parameters and proposes initial motion conditions,
such as object mobility, mass, friction, damping, and velocity. The Builder
validates object references, body types, collision assets, parameter
ranges, and required initial states, then initializes the simulation environment
by assembling the selected dynamic and static actors before preparing the
simulator command.
The language module therefore produces a structured plan, while deterministic
tools perform the actual execution. The validated scene is loaded into the
physics simulator with the same collision proxies used during reconstruction.
After simulation, the pose sequence is written back to the textured visual
meshes and rendered from the optimized original camera. Thus, simplified
collision geometry is used for contact and simulation, while detailed visual
geometry is kept for original view rendering.

To preserve image regions that are not represented by foreground meshes, we
maintain a separate static background used only for rendering. The system first
completes the background behind the segmented instances through mask guided
image editing and aligns the completed background with the optimized original
camera. Foreground meshes remain the only dynamic actors used for collision and
simulation; the background stays static and does not enter the scene graph,
collision tests, or physics solver. In the final rendering stage, the simulated
foreground meshes are composited with the aligned background from the optimized
original camera. This separation allows language modules to reason about roles,
causal phases, and physical assumptions, while deterministic tools handle plan
validation, geometry processing, simulation, and rendering.

\section{Experiments}
\label{sec:experiments}

We evaluate the proposed F3SR along three dimensions: agreement with the input
image, geometric contact quality of the generated scene, and the ability of
downstream agentic simulator assistance to plan, execute, and verify
physical goals.
We first describe the experimental setup and then present quantitative and
qualitative comparisons with scene generation methods, component ablations,
and a controlled ablation of agentic simulator assistance.

\subsection{Experimental Setup}

\noindent \textbf{Implementation Details.}
We evaluate on a curated set of single images containing multiple objects,
tabletop scenes, floor-level indoor scenes, and support heavy stacks.
Each sample contains an RGB image, instance masks, and a separate ground
prompt.
A scene goal subset pairs reconstructed scenes with falling, pushing,
dropping, and collision goals for the simulation agent study.
For internal variants, we keep the input masks, object assets, monocular
geometry, and the initial camera fixed.
All methods are rendered at the same resolution. F3SR uses the input
camera predicted and corrected by its full pipeline. When an external method
does not expose a compatible input view camera, we initialize its orientation
and intrinsics using the first camera alignment stage of our method. Starting
from this initialization, each external method independently optimizes a
bounded camera using mIoU between its own reconstructed
silhouettes and the corresponding input masks. The resulting method specific
camera is then fixed for all image metrics.
For penetration and stability evaluation, all methods are processed under the
same simulation settings. We release each final
scene without any task driven perturbation and keep only the ground plane and
background fixed. For physical parameter estimation in the agentic simulation stage, we use
OmniFysics~\cite{han2026omnifysics} to provide object-level physical priors,
including physical properties and dynamic conditions, which are then checked
against the allowed simulator parameter ranges before rollout execution.
The deterministic scene pipeline is run once per sample.
For the agentic ablation, each LLM assistance condition is evaluated once on
several fixed tasks across the reconstructed scenes. Retries are allowed only for communication interruptions during model calls.
Once a model response is obtained, malformed outputs, invalid simulation plans,
runner errors, and missing videos are counted as failures.
For qualitative original view rendering, we use a clean background
aligned with the optimized original camera.

\noindent \textbf{Baselines.}
The scene generation comparison includes three representative baselines with
different assumptions about image conditioned reconstruction and physical scene
assembly. SAM3D estimates per instance 3D objects from a single image
\cite{sam3dteam2025sam3d3dfyimages}. TabletopGen targets instance level
interactive tabletop scene generation from text or a single image
\cite{wang2025tabletopgen}. PAT3D is a physics augmented text based 3D scene
generation method, which we adapt to our single image setting by starting from
the same reconstructed scene assets before applying its physics oriented scene
perturbation protocol \cite{lin2025pat3d}.
For the simulator assistance ablation, GPT-5.5 and Qwen3.7-Plus receive
identical scene digests, images, task prompts, output constraints, and
deterministic execution pipelines. For each backend, we compare the full
Planner, Reasoner, and Builder workflow with a direct baseline that
predicts the final simulation plan without typed intermediate stages.

\noindent \textbf{Metrics.}
We use five metrics that jointly measure input view preservation, geometric
validity, and short and long horizon physical stability.
PSNR is computed from the method's final scene rendered with the
camera defined by the protocol above, before the standardized passive rollout. The
error is restricted to the union of input instance masks so that the
appearance only background does not dominate the score.
Image grounding is measured by mIoU, computed as the mean intersection over
union between each rendered instance silhouette and its corresponding input
mask.

Following PAT3D~\cite{lin2025pat3d}, we measure inter object penetration by
the penetration ratio $R_{\mathrm{pen}}$.
All objects are remeshed with the same target edge length relative to the
scene diagonal $\ell$. Let $N_{\mathrm{cross}}$ be the number of strict inter object triangle
crossings after remeshing, and let $\bar{\ell}_e$ be the mean remeshed edge
length. We define
\begin{equation}
  R_{\mathrm{pen}}
  =
  \frac{N_{\mathrm{cross}}\bar{\ell}_e}{\ell}.
  \label{eq:penetration-ratio}
\end{equation}
This dimensionless quantity approximates the total inter object intersection
contour length normalized by scene size and excludes self intersections inside
an individual generated mesh.

We
measure the maximum scene kinetic energy during the first second after release
and the maximum object drift over a 5 second passive rollout. This horizon
captures instability after release while remaining matched to the time scale
of our task level simulations. For dynamic object $i$, let $m_i$, $\mathbf v_i$,
$\mathbf I_i$, and $\boldsymbol\omega_i$ denote its mass, linear velocity,
world frame inertia, and angular velocity. The short horizon metric is
\begin{equation}
  K_{\max}
  =
  \max_{0\le t\le 1\,\mathrm{s}}
  \sum_i \left(
    \tfrac{1}{2}m_i\|\mathbf v_i^{(t)}\|_2^2
    + \tfrac{1}{2}{\boldsymbol\omega_i^{(t)}}^\top
      \mathbf I_i^{(t)}\boldsymbol\omega_i^{(t)}
  \right).
  \label{eq:max-kinetic-energy}
\end{equation}
Let $\mathbf c_i^{(t)}$ be object $i$'s center of mass. The rollout metric is
\begin{equation}
  D_{\max}
  =
  \max_i\max_{0\le t\le 5\,\mathrm{s}}
  \|\mathbf c_i^{(t)}-\mathbf c_i^{(0)}\|_2.
  \label{eq:max-drift-distance}
\end{equation}
We report $D_{\max}$ in centimeters after converting from the simulator
length unit.
Lower $R_{\mathrm{pen}}$, $K_{\max}$, and $D_{\max}$ indicate fewer geometric
intersections, less motion after release, and better rollout stability.

The agentic ablation reports three primary metrics. \emph{End to End Pass
Rate} is the fraction of the tasks that produce a valid final simulation
video. Provider side availability failures may be retried
until the model returns a response, after which the first infrastructure valid
outcome is retained. \emph{Prompt Following Rate} is measured
by manual binary audit over valid output videos. For each generated video, a
human reviewer assigns 1 if the result satisfies all task specific
requirements in the prompt and 0 otherwise; the reported rate is the mean of
these labels. The audit uses a
fixed task specific checklist for each prompt: the requested active object
must move, the motion must follow the specified direction or target object,
and the required event or outcome must be visibly realized. Stationary
objects, incorrect active objects, and wrong motions are failures.
\emph{Average End to End Time} is elapsed time from launching simulator
assistance until the final video is written, including LLM
inference, simulation, rendering, and background
compositing. Prompt following rates are computed over the valid output videos
from each condition.

\subsection{Quantitative Comparison}

We compare image conditioned scene generation methods under the same input,
rendering, and collision evaluation protocol.
PSNR and mIoU evaluate the reconstruction from the input view.
Penetration ratio measures residual inter object overlap that may be hidden by
occlusion, maximum kinetic energy detects unstable release or collision
divergence, and maximum drift distance measures stability over the passive
rollout.
All methods are aggregated over the same evaluation subset.

\begin{table}[!t]
  \caption{Main quantitative comparison on the evaluation subset.}
  \label{tab:main}
  \centering
  \resizebox{\linewidth}{!}{%
  \begin{tabular}{lccccc}
    \toprule
    Method & PSNR $\uparrow$ & mIoU $\uparrow$ & $R_{\mathrm{pen}}$ $\downarrow$ & $K_{\max}$ (J) $\downarrow$ & $D_{\max}$ (cm) $\downarrow$ \\
    \midrule
    SAM3D~\cite{sam3dteam2025sam3d3dfyimages} & 10.80 & \textbf{0.74} & 4.44 & $1.30{\times}10^{3}$ & $6.06{\times}10^{2}$ \\
    TabletopGen~\cite{wang2025tabletopgen} & 6.63 & 0.31 & \textbf{0.00} & 10.54 & 298.15 \\
    PAT3D~\cite{lin2025pat3d} & 7.13 & 0.47 & \textbf{0.00} & 3.13 & 5.31 \\
    \methodname{} & \textbf{11.39} & 0.71 & \textbf{0.00} & \textbf{0.04} & \textbf{1.09} \\
    \bottomrule
  \end{tabular}}
\end{table}

All metrics in Table~\ref{tab:main} are computed from the same output scene.
Jointly reporting input view scores, penetration, kinetic energy, and drift
exposes whether a method obtains visual agreement by accepting inter object
overlap or an unstable arrangement.
\methodname{} obtains the highest PSNR and the lowest release energy, while
SAM3D gives the highest mIoU. This difference reflects the expected
tradeoff between appearance-first reconstruction and simulation-ready scene
construction. Pixel PSNR and mIoU reward preserving the input view
projection, and a method can score well while leaving hidden inter-object
penetration or unstable support relationships in 3D. In contrast,
our method may move or separate objects during relation-aware contact
correction and physical settling, reducing some pixel-level agreement in order
to obtain a scene that can be released in the simulator. It eliminates measured
inter-object penetration on this subset and, relative to SAM3D, reduces release
energy by over four orders of magnitude and drift by over two orders of
magnitude. TabletopGen and PAT3D also reduce explicit penetration in the adapted
setting. The difference is that F3SR removes penetration while also
keeping a higher PSNR, lower release energy, and lower passive drift. In
particular, TabletopGen shows weak input view agreement and large passive drift,
and PAT3D is more stable than TabletopGen but still remains below F3SR
on PSNR, release energy, and drift. These results suggest that removing
penetration alone is insufficient for simulation-ready reconstruction.

\subsection{Qualitative Comparison}

\begin{figure}[!t]
  \centering
  \includegraphics[width=\linewidth]{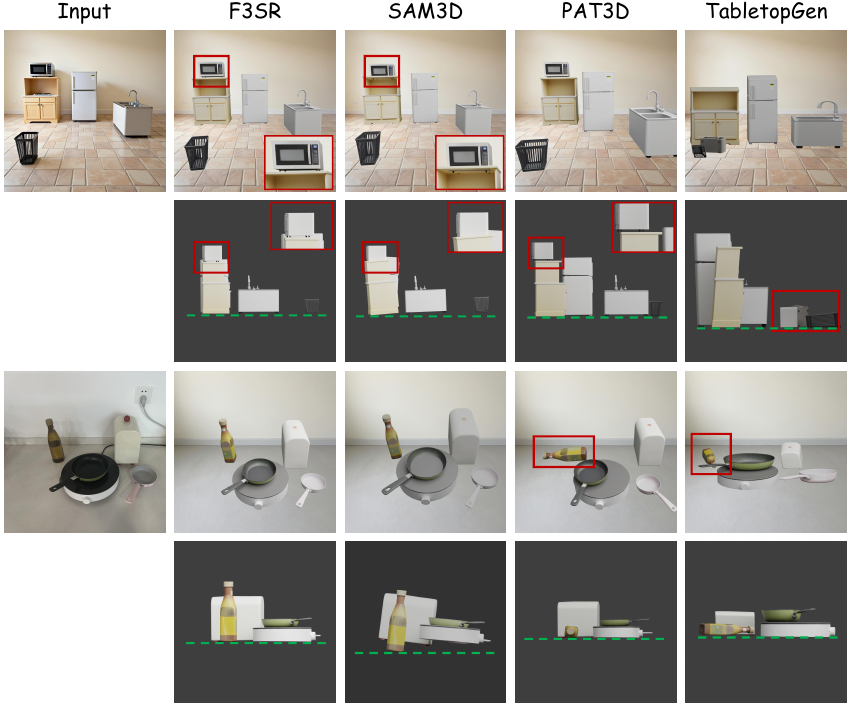}
  \caption{Qualitative comparison from the original and side views. Red boxes
  highlight representative differences or artifacts, and green lines mark the
  ground contact level for inspecting support, floating, and penetration.}
  \Description{Two qualitative examples with columns for the input image,
  \methodname{}, SAM3D, PAT3D, and TabletopGen. The examples include an indoor
  kitchen scene and a tabletop scene. Red boxes highlight local reconstruction
  failures or object layout inconsistencies.}
  \label{fig:qualitative-comparison}
\end{figure}

Figure~\ref{fig:qualitative-comparison} shows two representative examples
spanning a room-scale scene and a tabletop scene. For each input image, we
compare our F3SR with SAM3D, PAT3D, and TabletopGen using both the
original view and a side view. Since all object meshes in our pipeline are
initialized from SAM3D, we include SAM3D itself as a direct visual reference.
This comparison separates object-level reconstruction quality from scene-level
physical refinement: SAM3D can recover recognizable object assets, but the
assembled scene can still exhibit visible interpenetration, floating objects, or
unbalanced supports when the object meshes are placed without our grounding and
contact correction.

The side view provides a clearer and more intuitive inspection of simulation
readiness, including object support, vertical placement, and residual
interpenetration that may be hidden in the original image view. PAT3D produces physically motivated perturbations and can preserve a plausible
global layout in some cases, but its adapted results may still contain local
pose or placement errors that are inconsistent with the input image. TabletopGen explicitly
constructs object arrangements from a single image, but its tabletop-oriented
layout prior can still produce object placements and scale relationships that
deviate from the input view, especially when inspected from the side view.
In contrast, our method keeps the foreground
objects tied to the input view while correcting contacts and support relations,
which is reflected by fewer side view artifacts and more stable object-ground
alignment.

\subsection{Ablation Studies}

\noindent\textbf{Differentiable Rendering Alignment.}
The first ablation study removes the differentiable rendering update used to refine
each object's camera plane translation, $+Z$ yaw, and uniform scale. The
ablated variant keeps the same object meshes, initial camera, grounding,
contact correction, and passive settling stages as the full pipeline, so the
comparison isolates the effect of the per object alignment step. This step is
important because the initial object assets are reconstructed independently:
even when each mesh is visually plausible, small errors in image plane position,
scale, or yaw can accumulate into scene level layout errors after the objects
are assembled together.

\begin{figure}[!t]
  \centering
  \includegraphics[width=\linewidth]{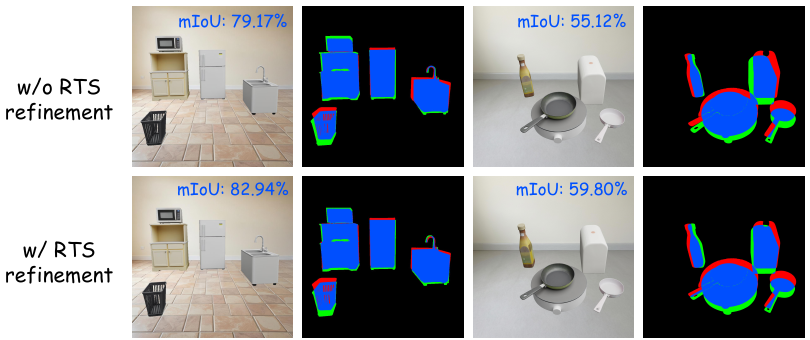}
  \caption{Visual ablation of differentiable rendering refinement. Red is the input mask, green is the
  rendered mask, and blue is their overlap.}
  \Description{Visual ablation of differentiable refinement. Columns show
  reconstruction without differentiable refinement, mIoU visualization without differentiable refinement, full
  reconstruction, and full mIoU visualization. Red pixels are ground truth mask
  regions, green pixels are reconstructed mask regions, blue pixels are overlap,
  and black pixels are background.}
  \label{fig:rts-ablation}
\end{figure}

Figure~\ref{fig:rts-ablation} visualizes this effect on representative scenes.
In the mIoU panels, red pixels denote the ground truth mask region, green pixels
denote the reconstructed mask region, blue pixels denote their overlap, and the
background is black. Without differentiable alignment, the rendered foreground
can remain close to the input in a coarse sense but miss object boundaries and
relative placements, especially for nearby objects whose masks constrain
support and contact. With alignment, the reconstructed masks overlap the input
masks more tightly, giving the following physical correction stage a better
initialized scene.

Removing differentiable rendering lowers PSNR from 11.39 to 11.29 dB and mIoU
from 0.71 to 0.67. The same ablation also increases the maximum release energy
from 0.04 J to 0.16 J and the maximum drift from 1.09 cm to 1.58 cm. This trend
suggests that input view alignment is not only a visual refinement step: it also
reduces the geometric inconsistency that contact correction and passive settling
must absorb. We therefore treat differentiable rendering as the bridge between
single image visual grounding and the later simulation-ready scene state.

\begin{table}[!t]
  \caption{Ablation of differentiable per object alignment.}
  \label{tab:ablation}
  \centering
  \resizebox{\linewidth}{!}{%
  \begin{tabular}{lccccc}
    \toprule
    Variant & PSNR $\uparrow$ & mIoU $\uparrow$ & $R_{\mathrm{pen}}$ $\downarrow$ & $K_{\max}$ (J) $\downarrow$ & $D_{\max}$ (cm) $\downarrow$ \\
    \midrule
    w/o differentiable rendering & 11.29 & 0.67 & \textbf{0.00} & 0.16 & 1.58 \\
    w/ differentiable rendering & \textbf{11.39} & \textbf{0.71} & \textbf{0.00} & \textbf{0.04} & \textbf{1.09} \\
    \bottomrule
  \end{tabular}}
\end{table}

\noindent\textbf{Damping for Gravity Settling.}
The second reconstruction ablation studies the damping used during passive
gravity settling. After relation aware contact correction, the scene can still
contain small residual torques from mesh roughness, imperfect supports, limited
collision proxy accuracy, or slightly tilted objects. These errors are often
small in the rendered image, but they become amplified once the scene is
released under gravity. Without damping, residual velocity and angular motion
can persist across the settling horizon, causing stacked objects to tip, slide
away from their supports, or trigger secondary contacts that are not implied by
the input image.

\begin{figure}[!t]
  \centering
  \includegraphics[width=\linewidth]{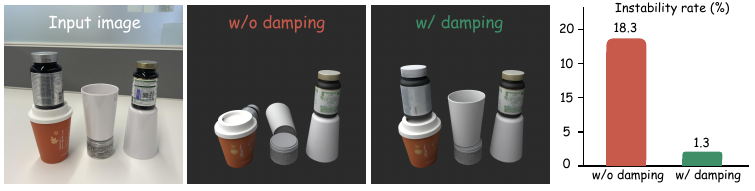}
  \caption{Ablation of damping during gravity settling. Damping suppresses
  residual release motion and lowers the probability of unstable object behavior.}
  \Description{Damping ablation with an input image, a reconstruction without
  damping where objects become unstable, a reconstruction with damping where
  objects remain supported, and a bar chart showing a lower unstable object
  ratio with damping.}
  \label{fig:damping-ablation}
\end{figure}

Damping is therefore used as a stabilization mechanism for reconstruction
release, rather than as a way to force task rollouts to succeed. It dissipates
the small residual motion introduced by imperfect single image reconstruction
while preserving the intended support layout. With damping enabled, the same
released scene converges to a stable configuration instead of drifting away from
the visually grounded initialization. Figure~\ref{fig:damping-ablation} shows
one representative example and the aggregate instability rate computed from 60
trials across several similar scenes. The instability rate is the percentage of
objects that topple, slide away from their support, or lose stable contact after
passive gravity settling.

\noindent\textbf{Agentic Simulator Assistance.}
We evaluate GPT-5.5 and Qwen3.7-Plus under the same task set, comparing each
backend with and without the three agentic roles. In the
full condition, the \emph{Planner} binds scene objects and produces causal
phases and success conditions; \emph{Reasoner} predicts semantic and physical
conditions; and \emph{Builder} validates object references, schemas, and
parameter ranges before invoking the fixed simulators. In the ablated
condition, the same LLM receives the same evidence and directly predicts the
final simulation plan schema in one pass, without role decomposition or typed
intermediate artifacts. The output is then passed to the same deterministic runner used by the full
agentic workflow.

All four conditions use the same reconstructed scenes, scene digests, crop
images, task prompts, parameter bounds, and simulator. The tasks cover toppling,
translation, directed contact, and launch and return events. Their fixed
checklists bind each prompt to the expected active object, target object when
present, motion direction, and visible outcome. This matched design tests
whether the staged roles affect operational reliability, visible prompt
following, and end to end cost independently of the chosen LLM backend.

\begin{table}[!t]
  \caption{Ablation of Planner, Reasoner, and Builder roles across LLM backends.}
  \label{tab:agentic-ablation}
  \centering
  \resizebox{\linewidth}{!}{%
  \begin{tabular}{llccc}
    \toprule
    LLM Backend & Simulator Assistance & E2E Pass (\%) $\uparrow$ & Prompt Following (\%) $\uparrow$ & Avg. E2E Time $\downarrow$ \\
    \midrule
    \multirow{2}{*}{GPT-5.5} & w/o Agentic Roles & \textbf{100.0} & 86.2 & 14.11 \\
    & w/ Agentic Roles & \textbf{100.0} & \textbf{93.1} & 13.44 \\
    \midrule
    \multirow{2}{*}{Qwen3.7-Plus} & w/o Agentic Roles & 93.1 & 0.0 & 12.95 \\
    & w/ Agentic Roles & 93.1 & 66.7 & \textbf{11.46} \\
    \bottomrule
  \end{tabular}}
\end{table}

Both GPT-5.5 settings produce a 100.0\% end to end pass rate, while both
Qwen3.7-Plus settings achieve 93.1\%. The remaining Qwen failures come from
invalid model outputs rather than the physics engine. Adding the agentic roles
improves prompt following from 86.2\% to 93.1\% for GPT-5.5 and from 0.0\% to
66.7\% for Qwen3.7-Plus. The staged workflow also reduces average end to end
time for both backends in this task set, from 14.11 to 13.44 seconds for
GPT-5.5 and from 12.95 to 11.46 seconds for Qwen3.7-Plus. We therefore
interpret the agentic roles as a structured prompt grounding mechanism that
improves visible task compliance while keeping runtime controlled.

\begin{figure}[!t]
  \centering
  \includegraphics[width=\linewidth]{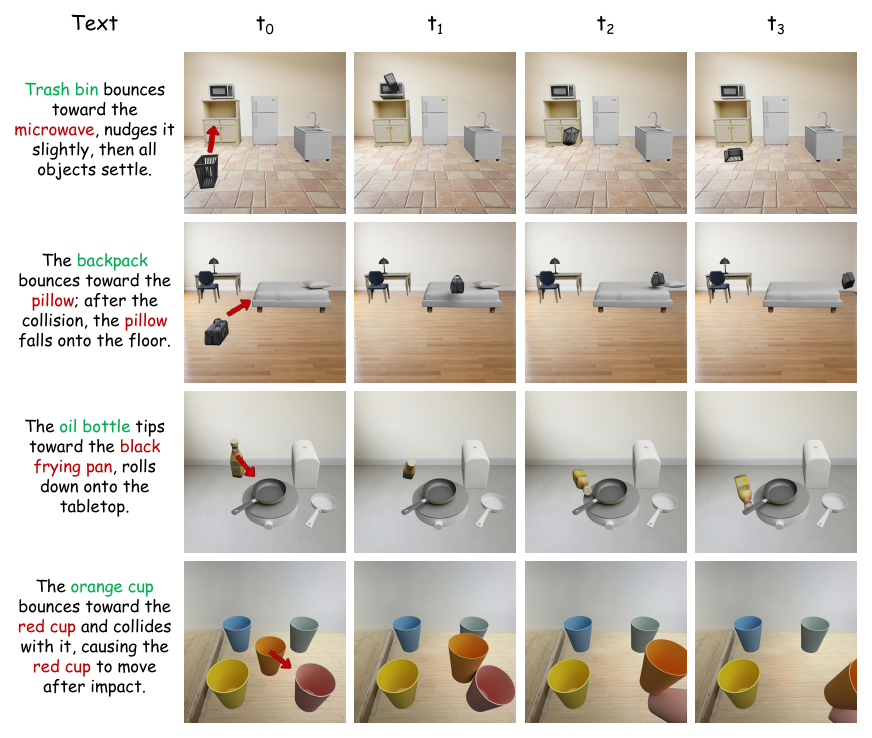}
  \caption{Goal conditioned \methodname{} simulation rollouts from the original
  view.}
  \Description{Four rows of \methodname{} simulation frames. The rows show a
  trash bin colliding with a microwave, a backpack colliding with a pillow, an
  oil bottle tipping toward a frying pan and rolling on the tabletop, and an
  orange cup colliding with a red cup so that the red cup moves after impact.}
  \label{fig:scene-gallery}
\end{figure}

\subsection{Goal Conditioned Simulation Rollouts}

Beyond static reconstruction and passive stability metrics, we further examine
whether the reconstructed scenes can support executable physical interactions.
This setting requires more than a visually plausible layout: objects must have
stable initial poses, usable collision geometry, consistent object identities,
and physical states that can be modified according to a task goal.

Given a simulation-ready scene and a scene specific physical goal, the agentic
simulation workflow first parses the goal into object roles and a short causal
plan. It identifies which object should be actuated, which object should respond,
and which surrounding objects should remain passive unless affected by contact.
The workflow then assigns the necessary physical controls, such as an initial
impulse or velocity, checks whether the involved objects are movable, and
verifies that the plan can be executed without changing the reconstructed visual
meshes. Invalid plans are rejected before simulation, while valid plans are
passed to the deterministic physics runner.

After execution, the simulated pose sequence is written back to the textured
visual meshes and rendered from the original camera view over the completed
clean background. This keeps the visualization tied to the input image while
allowing the scene to evolve under physical interaction. Figure~\ref{fig:scene-gallery}
shows representative rollouts across room scale and tabletop scenes. The
examples include impact driven motion, support changes, tipping and rolling,
and collision induced target motion. These results demonstrate that the
reconstructed scene is not only aligned for viewing, but also structured enough
for goal conditioned physical simulation.

\section{Conclusion}
\label{sec:discussion}

This work demonstrates that single-image 3D reconstruction can be extended from
visual asset recovery to simulation-ready scene construction when camera
grounding, gravity alignment, contact correction, and simulator execution are
treated as coupled scene-level requirements. \methodname{} keeps object meshes
tied to their input masks while refining their poses in a shared world frame and
converting the resulting scene into an executable physical environment. The
experiments show that strong visual reconstruction alone is not sufficient for
physical use: methods can preserve masks while leaving hidden instability, and
methods can remove penetration while still drifting or losing input-view
agreement. By combining grounded reconstruction with Planner, Reasoner, and
Builder roles for goal-conditioned simulation, the system provides a practical
bridge between pattern recognition, scene generation, and physical interaction.

\bibliographystyle{ieee_fullname}
\bibliography{refs}

\end{document}